\documentclass{article}

\usepackage{neurips_2022}

\usepackage[utf8]{inputenc} 
\usepackage[T1]{fontenc}    
\usepackage[hidelinks]{hyperref}       
\usepackage{url}            
\usepackage{booktabs}       
\usepackage{amsfonts}       
\usepackage{nicefrac}       
\usepackage{microtype}      
\usepackage{xcolor}         

\newcommand{\RR}{\mathbb{R}}

\newcommand{\N}{\mathbb{N}}

\newcommand{\M}{\mathcal{M}}
\newcommand{\OO}{\mathcal{O}}

\usepackage{amsmath}
\usepackage{amsthm}

\newtheorem{theorem}{Theorem}
\newtheorem{lemma}[theorem]{Lemma}
\newtheorem{corollary}[theorem]{Corollary}
\usepackage{mathtools}
\newtheorem{remark}{Remark}

\DeclarePairedDelimiter{\ceil}{\lceil}{\rceil}

\title{Expressivity of Shallow and Deep Neural Networks for Polynomial Approximation}

\author{%
  Itai Shapira\\
  Harvard University\\
}

\begin{document}

\maketitle

\begin{abstract}
    We delve into the required number of neurons for a Rectified Linear Unit (ReLU) neural network to approximate multivariate monomials. Our investigation establishes an exponential lower bound on the complexity of any shallow network approximating the product function $\textbf{x} \mapsto \prod_{i=1}^d x_i$ over a general compact domain. Moreover, we demonstrate this lower bound does not apply to normalized $\OO(1)$-Lipschitz monomials (or equivalently, by restricting to the unit cube). These results suggest that shallow ReLU networks suffer from the curse of dimensionality when expressing functions with a Lipschitz parameter scaling with the dimension of the input, and that the expressive power of neural networks lies in their depth rather than the overall complexity.
\end{abstract}

\section{Introduction}
Shallow neural networks, and by extension multi-hidden layer networks, are universal approximators. For every continuous non-polynomial activation function, the space of all shallow one-hidden layer networks is dense in $C(
K)$ in the uniform topology, for any compact $K \subset \RR^d$ (\cite{pinkus1999approximation}). Density, however, does not imply an efficient scheme for approximation, namely the required number of neurons and trainable parameters. Indeed, the minimum number of neurons required to $\varepsilon$-approximate a large class of functions can be exponential in the dimension of the input. Specifically, 
for the unit ball of the Sobolev space of order $r$ and dimension $d$,
\cite{maiorov1999approximation} show a lower bound of order $\OO(\varepsilon^{-\frac{d-1}{r}})$ and that the set of functions for which this lower bound holds is of large measure.

This lower bound reflects the so-called curse of dimensionality: the number of computational units necessary for approximation scales exponentially with the problem dimension. In this regard, shallow networks offer no superior order of approximation compared to other functional approximation schemes, such as polynomial approximation (\cite{pinkus1999approximation}).

In contrast, \textit{deep} networks featuring multiple hidden layers have proven effective in high-dimensional applications such as image recognition and natural language processing. In light of this empirical success, numerous studies have examined the approximation capabilities of deep versus shallow networks. It has been suggested that the expressive power of neural networks is attributed to their depth. When compared to shallow networks of equivalent size, deeper networks may offer greater expressivity and can efficiently capture functions that would demand exponentially-wide shallow networks (refer to section \ref{RelatedWork} for a review).
Possibly the most striking illustration of this is the \textit{depth separation phenomena}: studies such as \cite{eldan2016power}, \cite{safran2017depth}, \cite{daniely2017depth}, and \cite{osti_10329461} demonstrate that certain functions can be efficiently represented using depth two networks, yet require exponentially wider networks for approximation by shallow, one-hidden-layer networks.

A particularly interesting case study for the complexity gap between deep and shallow neural networks involves the class of homogeneous multivariate polynomials of $d$ variables, represented by the monomials:
\begin{align*}
p_d(x_1,..,x_d) := \prod_{i=1}^d x_i
\end{align*}
i.e. the $d$-product function. 
Deep networks can approximate $p_d$ efficiently.
Monomials are \textit{compositionally sparse}, implying they can be expressed by recursive compositions of the low-dimension function $(x,y) \mapsto xy$. Consequently, a tree-like network architecture with $\log d$ layers can approximate $p_d$ with a linear number of neurons (\cite{mhaskar2016learning}, \cite{poggio2017and}). This result prompts the following question:

\begin{quote}
\emph{Can an $\OO(1)$-layer network approximate $\textbf{x} \mapsto \prod_{i=1}^d x_i$ over a general compact domain $[-k,k]^d$ with only poly$(d)$ neurons?}
\end{quote}

\cite{lin2017does} and \cite{rolnick2017power} 
studied this problem in the context of \textit{exact approximation}, 
that is, how many neurons are necessary to have the property that for any precision $\varepsilon>0$, there exists a weight assignment that approximate $p_d$. For this notion of approximation, they demonstrated that an exponential number of neurons is needed for a shallow network with a smooth activation function. However, it remained uncertain whether a smaller network could be constructed to approximate $p_d$ arbitrarily well in the standard notion of approximation, where the number of neurons is allowed to vary with the level of accuracy $\varepsilon$.

\cite{blanchard2021shallow} recently constructed a two-layer ReLU network with poly$(d)$ neurons that approximates the product function over $[0,1]^d$. This corresponds to the normalized monomial $\textbf{x} \mapsto \frac{1}{k^d} \prod_{i=1}^d x_i$ over $[0,k]^d$, where $k$ is a constant. However, we demonstrate that the choice of approximation domain $[0,1]^d$ (or equivalently, the normalization factor $k^{-d}$) hides a gap between the positive and negative answers to the question of the expressive power of ReLU networks in the context of homogeneous polynomials.

In this paper, we offer a characterization of the expressive power of ReLU networks for homogeneous polynomials. We demonstrate that over the domain $(1,k]^d$, the product function acts as an expansive map with an expansion constant that scales exponentially with the dimension. As a result, we prove that no $\OO(1)$-layer ReLU network exists that can approximate $p_d$ with at most poly$(d)$ neurons.
Conversely, for the normalized case where the Lipschitz parameter is independent of the dimension, we extend the result of \cite{blanchard2021shallow}. We show that $p_d$ can be efficiently approximated by a one-hidden-layer shallow network with a smooth or ReLU activation function. In particular, we reveal that no depth separation exists between one- and two-hidden-layer networks when approximating the class of homogeneous polynomials.

Our main contributions can be summarized as follows:
\begin{itemize}
\item In Section \ref{section3}, we demonstrate that the minimum number of neurons required by any ReLU $\OO(1)$-layer network to approximate multivariate monomials \textbf{scales exponentially with the dimension} over $[-k,k]^d$ with $k > 1$. More specifically, we prove the following theorem:
\begin{theorem}
If a ReLU network with $L$ layers and at most $n$ neurons in each layer $\varepsilon$-approximates $\textbf{x} \mapsto \prod_{i=1}^d x_i$ over $[
-k,k]^d$ for $k>1$, then
$n = \exp (\OO( \frac{d}{L} \ln (\varepsilon^{-1})))$. (See Theorem \ref{mr1} for a formal statement)
\end{theorem}
\item In Section \ref{section5}, we demonstrate that the normalized monomial $\textbf{x} \mapsto k^{-d}\prod_{i=1}^d x_i$ over $[0,k]^d$ (equivalently, $p_d$ over $[0,1]^d$)
can be \textbf{approximated using a one-hidden layer ReLU network with poly$(d)$ neurons }
\end{itemize}

\section{Preliminaries}
\subsection{Feedforward Neural Networks}
Let $K \subset \RR^d$ be a compact set and $(C(K), || \cdot ||_\infty)$ denote the space of all continuous functions on $K$, equipped with the uniform norm: $||f|| = \max_{x \in K} |f(x)|$.
In this work, we contemplate the standard model of feedforward neural networks, using linear output neurons and a non-linear continuous activation function $\sigma: \RR \longrightarrow \RR$ for the other neurons. 
Following the notation in \cite{pinkus1999approximation},
we denote by $\M_n^1(\sigma)$ the set of all 1-hidden layer neural networks:
\begin{align*}
\M_n^1(\sigma) = \bigg\{ \sum_{i=1}^n \nu_i \sigma( \textbf{w}_i^T \textbf{x} + b_i ) \mid \nu_i, b_i \in \RR, \textbf{w}_i \in \RR^d 
\bigg\}
\end{align*}
Throughout this work, we adopt the convention of referring to $f \in \M_n^1(\sigma)$ as \textit{shallow networks}. For brevity, we also employ matrix notation $\M_n^1(\sigma) = \textbf{A}_1 \sigma(\textbf{A}_0\textbf{x}+\textbf{b}_0)$,
where $\textbf{A}_0$ is an $n \times d$ matrix, $\textbf{A}_1$ is a $1 \times n$ matrix and $\sigma$ is applied element-wise to vectors.
The total number of trainable parameters is $(d+2) n$.
A deep neural network with $L$ hidden layers is obtained by feeding the outputs of a given layer as inputs to the next:
\begin{align*}
\M_n^L(\sigma) =  \bigg\{ 
\textbf{A}_L(\sigma \cdots \sigma(\textbf{A}_1(\sigma(\textbf{A}_0 \textbf{x} + \textbf{b}_0) )+\textbf{b}_1)\cdots )
\bigg\} 
\end{align*}
where $n$ is the maximum number of neurons in each hidden layer.
The constant $L$ is referred to as the \textit{depth} of the network. 

We will consider two types of activation functions: piece-wise linear activation, such as the popular rectified linear unit (ReLU) $\sigma(x) := \max \{ x, 0 \}$; and general smooth non-polynomial activation functions, such as the exponential function $\exp(x) := e^x$.

\subsection{Approximation Complexity}
We will consider $L^\infty$-error of approximation. We say that a network $g \in \M_n^L(\sigma)$ $\varepsilon$-approximates a function $f$ if $|| f -g|| < \varepsilon$ in the uniform topology.
We measure the complexity of the network by the number of neurons and non-zero weights in the network. We are interested in evaluating the number of neurons needed to approximate a given function within $\varepsilon$, and especially how this number scales with the dimension of the problem $d$ and with the accuracy level $\varepsilon$. 

Roughly speaking, we consider shallow networks to be inefficient in approximating a sequence of functions $(f_d: \RR^d \longrightarrow \RR )_{d \in \N}$, if the minimum number of neurons needed to $\varepsilon$-approximate the sequence grows exponentially with $d$. Conversely, for any fixed $\varepsilon $, polynomial dependency on $d$ is considered efficient. Note, however, that we do not require a polynomial dependence on $\varepsilon^{-1}$ and $d$ simultaneously.
For the remainder of the paper, we will use $\OO$ notation which hides constants independent of $d$.

\section{Related Work} \label{RelatedWork}

\textbf{Slow approximation by shallow networks in standard function spaces.} Several studies have shown that shallow networks are inefficient in approximating Sobolev functions.
\citet{maiorov1999approximation} consider the rate of approximation by arbitrary ridge functions. They show a lower bound that scales exponentially with $d$, thus demonstrating the inherent inefficiency of \textit{any} shallow-like approximation schemes.
\citet{devore1989optimal} proved that any continuous function approximator that $\varepsilon$-approximates functions from the unit ball in the Sobolev space of order $r$ and dimension $d$ needs at least $\Theta(\varepsilon^{-\frac{d}{r}})$ parameters (note, however, that the optimal weight selection is generally not continuous). \citet{pinkus1999approximation} and \citet{yarotsky2017error} prove the existence of norm-one Sobolev functions that cannot be approximated efficiently by shallow network, for smooth and ReLU activation functions, respectively.

\textbf{Faster approximation by deep networks.} Several studies have shown that deeper networks perform better for a given number of neurons. This indicates that the expressive power of neural networks lies in their depth rather than the overall complexity.
\citet{telgarsky2015representation} show a $k$-layer $\OO(1)$-wide 1-dimensional ReLU network which oscillates $\OO(2^k)$ times that cannot be approximated by a $k$-polynomial shallow network. \citet{yarotsky2017error} show every $f \in C^2([0,1]^d)$ cannot be $\varepsilon$-approximated by a $L$-deep ReLU network with fewer than $\OO(\varepsilon^{-\frac{1}{2L}})$ neurons, demonstrating the efficiency of increased depth. \citet{mhaskar2016learning} and \citet{poggio2017and} show that deep neural networks can efficiently express \textit{compositionally sparse} functions, i.e. functions that can be expressed by recursive compositions of low-dimension functions. Other authors considered the power of deeper networks of different types. The exponential benefit of depth was shown by \citet{delalleau2011shallow} (networks consisting of sum and product nodes) and \citet{cohen2016expressive} (convolutional arithmetic circuit architecture that incorporates locality, sharing, and pooling).

\textbf{Separation gaps.}
Several works have studied the gap in expressivity between one-hidden-layer and two-hidden-layer networks, and have proved the existence of functions that can be efficiently approximated by two-hidden-layer networks, but require exponential width shallow networks. \citet{eldan2016power} prove a separation gap for rapidly oscillating radial functions using ReLU networks. Similar separation gaps have been shown by \citet{daniely2017depth}, \citet{safran2017depth}, and \citet{osti_10329461}.

\textbf{Flattening results.}
Recent work studied the complexity cost of flattening deep networks into shallow ones.
\citet{safran2019depth} discuss flattening networks which approximate $\OO(1)$-Lipschitz radial functions.
\citet{osti_10329461} show that functions with an $\OO(1)$-rate of oscillation can be approximated by one-hidden-layer networks.

\textbf{Approximation of the product function.}
The product function is a special case of compositionally-sparse functions and thus can be approximated by deep networks, as shown in \citet{mhaskar2016learning} and \citet{poggio2017and}. \citet{lin2017does} and \citet{rolnick2017power} studied the \textit{exact}-approximation capabilities of shallow and deep networks in approximating this function. They proved that if $m(\varepsilon)$ is the minimum number of neurons required by a smooth shallow network to $\varepsilon$-approximate $p_d$, then $\lim_{\varepsilon \to 0} m(\varepsilon)$ exists and equals to $2^d$ (In Appendix \ref{new_proof_to_Lin}, we attached a slightly shorter proof). More recently, \citet{blanchard2021shallow} constructed a two-hidden-layer ReLU architecture that $\varepsilon$-approximates the normalized $p_d$ with $\OO(d^{\frac{3}{2}} \varepsilon^{-\frac{1}{2}} \ln \varepsilon^{-1})$ neurons.

\section{The Inefficiency of Shallow Network on a General Compact Domain} \label{section3}
In this section we provide an exponential lower bound on the complexity of a ReLU network that $\varepsilon$-approximate the multivariate monomial by counting the number of linear regions in which the network is linear. 

Shallow networks with a piecewise linear activation function compute piecewise linear functions.
ReLU activation functions, of the form $\max \{ \textbf{w}^T \textbf{x} + b , 0 \}$, 
operate in one of two modes - they either output $0$ or a linear function of the input. The boundary between these two behaviors is the hyperplane $H=\{ \textbf{x} \mid \textbf{w}^T \textbf{x} + b = 0 \}$ which splits the input space $\RR^d$ into two pieces. A shallow network with $n$ neurons forms an $d$-dimensional \textit{hyperplane arrangement}   $\{ H_i \subset \RR^d \}_{i \in [n]}$. A  \textit{linear region}  of an arrangement
is a connected component of $\RR^d \setminus \bigcup_{i \in [n]} H_i$ 
(\cite{pascanu2013number}). In every such region, the inference function of the network is affine linear.

An arbitrary non-linear function, when defined on a sufficiently large set, cannot be approximated by linear functions. For the sake of building intuition for the below result, consider the following example inspired by \cite{liang2016deep} (Theorem 11).
Suppose that $f: \RR^d \longrightarrow \RR$ is strongly-convex with parameter $m>0$ and $T$ is a linear function that $\varepsilon$-approximates $f$ over the domain $K$.
Then the error function $g(x):= f(x) - T(x)$ is also strongly convex. 
If $\textbf{x}_0 \in \RR^d$ is any point with $\nabla g(\textbf{x}_0) = 0$, then by definition:
\begin{align*}
2\varepsilon >  g(\textbf{y}) - g(\textbf{x}_0) = \frac{m}{2} || \textbf{y}-\textbf{x}_0||_2^2
\end{align*}
that is, in order for $T$ to $\varepsilon$-approximates $f$, the domain $K$ must have a relatively small diameter. By finding an upper bound on the number of linear regions generated by the network, we can derive a lower bound on the number of neurons required to approximate a general (not necessarily strongly convex) function $f$.

The topic of counting the number of linear regions generated by a ReLU network has been addressed by several authors (\cite{montufar_number_2014},\cite{telgarsky2015representation},\cite{pascanu2013number}. See also \cite{zaslavsky1975facing}). The function represented by the network can have a number of linear pieces that is exponential in the number of layers, but is at most polynomial in the number of neurons
 (\cite{hanin2019complexity}). An upper bound for the case of $d=1$ is given by
\cite{telgarsky2015representation} (see also \cite{yarotsky2017error} (Lemma 4)):
\begin{lemma}[\cite{telgarsky2015representation} (Lemma 2.1)] \label{regions}
Consider $f \in \M_n^L(\sigma)$, a $1$-dimensional ReLU network with $L$ hidden layers and no more than $n$ neurons in each hidden layer. The number of linear pieces in $f$ is at most $(2n)^L$.
\end{lemma}

Using Lemma \ref{regions}, we can now prove the main result of this section:
\begin{theorem} \label{mr1}
If $k>1$ and $\varepsilon > 0$, then any network $f \in \M_n^L(\text{ReLU})$ that $\varepsilon$-approximates the univariate monomial $p_d(x) = \prod_{i=1}^d x_i$ on $K:=[-k,k]^d$, must satisfy $n>C_1 \varepsilon^{- \frac{1}{2L}} d^{\frac{1}{2L}} \exp (C_2\frac{d}{L} )$, where $C_1,C_2>0$ are constants independent of $d$ and $\varepsilon$. Specifically,
$n$ must scale exponentially with $d$.
\end{theorem}

\begin{proof}
The shallow network $f$ $\varepsilon$-approximates $p_d$ in $K$ and, in particular, approximates $p$ along the direction $\lambda (1,1,..,1)$.
We define $\tilde{p}_d, \tilde{f}: [\frac{k+1}{2}, k] \longrightarrow \RR$ as the restrictions of $p_d$ and $f$ respectively along this direction:
\begin{align*}
\tilde{f}(x) = f(x,...,x) \quad \tilde{p}_d(x) = x^d
\end{align*}
clearly $|\tilde{f}(x) - \tilde{p}_d(x)| < \varepsilon$ for every $x \in [\frac{k+1}{2}, k]$. Additionally, $\tilde{f}$ is a univariate shallow network with $n$ neurons, obtainable from $f$ by replacing the $d$ input units with one input and modify the connection accordingly.

 $\tilde{f}$ is a continuous piece-wise affine linear function with $(2n)^L$ linear pieces. Therefore, the domain $[\frac{k+1}{2}, k]$ is partitioned into at most $(2n)^L$ intervals for which $\tilde{f}$ is linear. Hence, there exists an interval $[a,b] \subset [\frac{k+1}{2}, k]$ with $b-a \geq \frac{k-1}{2(2n)^L}$ such that $\tilde{f}$ is linear on $[a,b]$ and $\varepsilon$-approximates $x^d$. Denote by $\tilde{h}$ the error function $\tilde{h}:=(\tilde{p}_d(x)-\tilde{f}(x) )$. Notice that: $|\tilde{h}(x)| < \varepsilon$ for all $x \in [a,b]$. 
 
Consider the following three points $a,b$ and $\frac{a+b}{2} \in [a,b]$. It follows from the linearity of $\tilde{f}$:
\begin{align} \label{rate_of_change}
\begin{split}
4 \varepsilon > \tilde{h}(b) + \tilde{h}(a) - 2\tilde{h}(\frac{a+b}{2}) &= \bigg(\tilde{p}_d(b)-\tilde{p}_d(\frac{a+b}{2}) \bigg)+\bigg(\tilde{p}_d(a)-\tilde{p}_d(\frac{a+b}{2}) \bigg) \\
&+\bigg(\tilde{f}(\frac{a+b}{2})-\tilde{f}(b) \bigg)+\bigg(\tilde{f}(\frac{a+b}{2})-\tilde{f}(a) \bigg) \\
&=\bigg(\tilde{p}_d(b)-\tilde{p}_d(\frac{a+b}{2}) \bigg)+\bigg(\tilde{p}_d(a)-\tilde{p}_d(\frac{a+b}{2}) \bigg) 
\end{split}
\end{align}
Notice that on $[\frac{K+1}{2}, K]$, the function $\tilde{p}_d$ is strongly convex with parameter $m:= d(d-1)(\frac{k+1}{2})^{d-2}$. It follows than:
\begin{align*}
       \tilde{p}_d(b) +  \tilde{p}_d(a)- 2\tilde{p}_d( \frac{a+b}{2} ) \geq \frac{m}{4} (b-a)^2 
\end{align*}

It follows that: $4 \varepsilon > \frac{m}{4}(b-a)^2$ and using the fact that $b-a \geq \frac{k-1}{2(2n)^L}$:
\begin{align*}
  4 \varepsilon > \frac{d(d-1)}{16} (\frac{k+1}{2})^{d-2} (k-1)^2 (2n)^{-2L}
\end{align*}
that is, exists $C_1,C_2 > 0$ independent of $d$ and $\varepsilon$ with: $n > C_1 \varepsilon^{- \frac{1}{2L}} d^{\frac{1}{2L}} \exp (C_2\frac{d}{L} )$.

\end{proof}

\subsection*{Discussion}
Theorem \ref{mr1} tells us that $\varepsilon$-approximating the product function outside the unit cube requires exponentially wide $\OO(1)$-deep ReLU networks. The proof relies on the fact that $\textbf{x} \mapsto \prod_{i=1}^d x_i$ is expansive, with an expansion factor that scales exponentially with $d$. 
In contrast, over $[0,1]^d$, $p_d$ is $1$-Lipschitz, and this Lipschitz parameter does not scale with $d$. Note that expressing $p_d$ over this domain
is equivalent to approximate the \textit{normalized monomial}:
\begin{align} \label{normalized_monomial}
\textbf{x} \in [0,k]^d \mapsto \frac{1}{k^d} \prod_{i=1}^d x_i
\end{align}
Recently, \cite{blanchard2021shallow} showed the following:
\begin{theorem}[\cite{blanchard2021shallow}, Proposition 3.2] \label{two-layers-mono}
For all $\varepsilon > 0$, there exists a two-hidden-layer ReLU network with poly$(d)$ neurons that $\varepsilon$-approximates $\textbf{x} \in [0,1]^d \mapsto \prod_{i=1}^d x_i$
\end{theorem} 
Together, Theorems \ref{mr1} and \ref{two-layers-mono}  suggest that 
the curse of dimensionality of shallow ReLU networks is closely related to the derivative of the objective function and how its magnitude scales with the dimension.
Analogously, in a seminal work, \cite{eldan2016power} and 
\cite{safran2019depth}
investigated radial functions of the form $f(\textbf{x}) = \varphi(||\textbf{x}||)$. The former demonstrated that no one-hidden layer network can approximate $f$ if $\varphi$ is rapidly oscillating, with a Lipschitz parameter scaling polynomially with $d$. Meanwhile, the latter established that if $\varphi$ is $\OO(1)$-Lipschitz, approximation by a shallow network is feasible with poly$(d)$ neurons. 
In our framework, the proof of \ref{two-layers-mono} relies on the observation that $p_d(\textbf{x}) = \exp(\sum_{i=1}^d \ln x_i)$ and that $\exp$ is $\OO(1)$-Lipschitz if $x_i \leq 1$.

\section{Fast Approximation Using Shallow Networks For Normalized Monomials} \label{section5}
Extending the findings of \cite{blanchard2021shallow}, in this section we raise the following question:
\begin{quote}
\emph{Is there a depth separation between one and two-hidden-layer networks in the approximation of the normalized monomial?}
\end{quote}
In other words, does a shallow counterpart to Theorem \ref{two-layers-mono} with poly$(d)$ number of neurons exist?
In the following theorem, we address this question by constructing a shallow ReLU network with poly$(d)$ number of neurons that $\varepsilon$-approximates the normalized $p_d$.

\begin{theorem} \label{mr2}
Let $k >0$. For any $\varepsilon > 0$, there exists a shallow network $\hat{f} \in \M_n^1(\text{ReLU})$ that $\varepsilon$-approximates $p_d:[0,k]^d \longrightarrow \RR$, defined by $p_d(\textbf{x}) = \frac{1}{k^d} \prod_{i=1}^d x_i$, with $n = \text{poly}(d)$.
\end{theorem}
The full proof can be found in Appendix \ref{proof}. 
This proof is constructive and is built upon the following observation: a polynomial $q_r$ of degree $r$, which depends solely on $\varepsilon^{-1}$, exists that $\varepsilon$-approximates the univariate function $x \mapsto e^x$. Thus, we have:
\begin{align} \label{idea}
p_d(\textbf{x}) = \exp\bigg( \sum_{i=1}^d \ln x_i \bigg) \approx q_r\bigg( \sum_{i=1}^d \ln x_i \bigg)
\end{align}
which reduces the problem to approximating polynomials of degree at most $r$ in the $d$ variables $(\ln x_i)_{i=1}^d$.
Observe that there are only $\binom{d+r}{r} \leq (d+1)^r$ elements in the right-hand side of equation (\ref{idea}). This observation enables us to "flatten" the network in Theorem \ref{two-layers-mono} into a one-hidden-layer network. 
As suggested by Theorem \ref{mr1}, the normalization factor in equation (\ref{normalized_monomial}) is critical for the proof.
Similar flattening results are shown in \cite{safran2019depth} Theorem 1 (for Lipschitz radial functions) and in the more general \cite{osti_10329461} Theorem 11 (using Fourier networks). The crux of the proof is demonstrating that the product function meets the conditions of the latter result, with the remaining details presented in a simplified form for completeness.

\textbf{Proof Sketch.}  Essentially, we construct a two-layer network akin to Theorem \ref{two-layers-mono} with the $\exp$ activation function and then employ the framework previously described to demonstrate that flattening to a shallow network is feasible with a polynomial cost in complexity. 

The high-level strategy for this proof proceeds in several key stages. Initially, a two-hidden-layer network is constructed with the $\exp: x \mapsto e^x$ activation function, such that $\hat{f}(\textbf{x}) = \exp \bigg( \sum_{i=1}^n \nu_i \exp(\textbf{w}_i \textbf{x}) \bigg)$ approximates $p_d$. Additionally, it is demonstrated that it is feasible to control the value of the coefficients within the network. 

Following this, the proof establishes that the non-linearity of the second hidden layer can be approximated by a polynomial with a linear cost in $\varepsilon^{-1}$. For a specific univariate polynomial $q$, whose degree depends solely on $\varepsilon$, $\hat{f}$ is substituted by 
\begin{align} \label{poly-on-net}
q\bigg( \sum_{i=1}^n \nu_i \exp(\textbf{w}_i \textbf{x})\bigg)
\end{align}
It is important to note that the normalization factor in (\ref{normalized_monomial}) is essential for this stage of the proof.
We then leverages properties of the exponential function, specifically the fact that $\exp(\textbf{w} \textbf{x})^m = \exp((m \textbf{w})\textbf{x})$, to demonstrate that (\ref{poly-on-net}) is a shallow $\exp$ network. An analogous result for shallow Fourier neural networks is presented in \cite{osti_10329461} Lemma 33. 
The proof is completed by demonstrating that the $\exp$ activation function in the shallow network can be replaced with the ReLU activation function.

\section{Conclusion}
We have established results describing the expressive power of $\OO(1)$-ReLU-networks in the context of approximating the class of homogeneous multivariate polynomials. 

\textbf{Deep vs shallow}. 
Our investigation provides further evidence that deep ReLU networks demonstrate superior efficiency in expressing homogeneous polynomials. The number of computational units necessary for expressing the product function diminishes significantly with increased depth, as evidenced by the lower bound in Theorem \ref{mr1}. With $L=\log d$ layers, a deep network can efficiently express this function.

\textbf{The curse of dimensionality.} 
Our findings suggest that the product function can be efficiently expressed using a neural network if the network is sufficiently deep to exploit the computational structure of the function (\cite{poggio2017and}), or if it operates on a domain in which its Lipschitz constant does not grow with the dimension. This observation aligns with the surprising recent result in \cite{safran2019depth}. In their respective works, \cite{daniely2017depth} and \cite{eldan2016power} demonstrated that functions of the form $x \mapsto \varphi(||x||)$ can be approximated by depth-two networks, leveraging the computational structure. However, unless $\varphi$ is $\OO(1)$-Lipschitz (\cite{safran2019depth}), it cannot be expressed efficiently using a one-hidden-layer network.

\newpage

\bibliographystyle{plainnat} 
\bibliography{reference} 

\begin{thebibliography}{26}
\providecommand{\natexlab}[1]{#1}
\providecommand{\url}[1]{\texttt{#1}}
\expandafter\ifx\csname urlstyle\endcsname\relax
  \providecommand{\doi}[1]{doi: #1}\else
  \providecommand{\doi}{doi: \begingroup \urlstyle{rm}\Url}\fi

\bibitem[Ash(1970)]{ash1970characterization}
J~Marshall Ash.
\newblock A characterization of the peano derivative.
\newblock \emph{Transactions of the American Mathematical Society},
  149\penalty0 (2):\penalty0 489--501, 1970.

\bibitem[Blanchard and Bennouna(2021)]{blanchard2021shallow}
Moise Blanchard and Mohammed~Amine Bennouna.
\newblock Shallow and deep networks are near-optimal approximators of korobov
  functions.
\newblock In \emph{International Conference on Learning Representations}, 2021.

\bibitem[Cohen et~al.(2016)Cohen, Sharir, and Shashua]{cohen2016expressive}
Nadav Cohen, Or~Sharir, and Amnon Shashua.
\newblock On the expressive power of deep learning: A tensor analysis.
\newblock In \emph{Conference on learning theory}, pages 698--728. PMLR, 2016.

\bibitem[Corominas and Balaguer(1954)]{corominas1954condiciones}
Ernesto Corominas and Ferran~Sunyer Balaguer.
\newblock Condiciones para que una funcion infinitamente derivable sea un
  polinomio.
\newblock \emph{Revista matem{\'a}tica hispanoamericana}, 14\penalty0
  (1):\penalty0 26--43, 1954.

\bibitem[Daniely(2017)]{daniely2017depth}
Amit Daniely.
\newblock Depth separation for neural networks.
\newblock In \emph{Conference on Learning Theory}, pages 690--696. PMLR, 2017.

\bibitem[Delalleau and Bengio(2011)]{delalleau2011shallow}
Olivier Delalleau and Yoshua Bengio.
\newblock Shallow vs. deep sum-product networks.
\newblock \emph{Advances in neural information processing systems}, 24, 2011.

\bibitem[DeVore et~al.(1989)DeVore, Howard, and Micchelli]{devore1989optimal}
Ronald~A DeVore, Ralph Howard, and Charles Micchelli.
\newblock Optimal nonlinear approximation.
\newblock \emph{Manuscripta mathematica}, 63\penalty0 (4):\penalty0 469--478,
  1989.

\bibitem[Eldan and Shamir(2016)]{eldan2016power}
Ronen Eldan and Ohad Shamir.
\newblock The power of depth for feedforward neural networks.
\newblock In \emph{Conference on learning theory}, pages 907--940. PMLR, 2016.

\bibitem[Hanin and Rolnick(2019)]{hanin2019complexity}
Boris Hanin and David Rolnick.
\newblock Complexity of linear regions in deep networks.
\newblock In \emph{International Conference on Machine Learning}, pages
  2596--2604. PMLR, 2019.

\bibitem[Leshno et~al.(1993)Leshno, Lin, Pinkus, and
  Schocken]{leshno1993multilayer}
Moshe Leshno, Vladimir~Ya Lin, Allan Pinkus, and Shimon Schocken.
\newblock Multilayer feedforward networks with a nonpolynomial activation
  function can approximate any function.
\newblock \emph{Neural networks}, 6\penalty0 (6):\penalty0 861--867, 1993.

\bibitem[Liang and Srikant(2016)]{liang2016deep}
Shiyu Liang and Rayadurgam Srikant.
\newblock Why deep neural networks for function approximation?
\newblock \emph{arXiv preprint arXiv:1610.04161}, 2016.

\bibitem[Lin et~al.(2017)Lin, Tegmark, and Rolnick]{lin2017does}
Henry~W Lin, Max Tegmark, and David Rolnick.
\newblock Why does deep and cheap learning work so well?
\newblock \emph{Journal of Statistical Physics}, 168\penalty0 (6):\penalty0
  1223--1247, 2017.

\bibitem[Maiorov et~al.(1999)Maiorov, Meir, and
  Ratsaby]{maiorov1999approximation}
Vitaly Maiorov, Ron Meir, and Joel Ratsaby.
\newblock On the approximation of functional classes equipped with a uniform
  measure using ridge functions.
\newblock \emph{Journal of approximation theory}, 99\penalty0 (1):\penalty0
  95--111, 1999.

\bibitem[Mhaskar et~al.(2016)Mhaskar, Liao, and Poggio]{mhaskar2016learning}
Hrushikesh Mhaskar, Qianli Liao, and Tomaso Poggio.
\newblock Learning functions: when is deep better than shallow.
\newblock \emph{arXiv preprint arXiv:1603.00988}, 2016.

\bibitem[Mhaskar(1996)]{mhaskar1996neural}
Hrushikesh~N Mhaskar.
\newblock Neural networks for optimal approximation of smooth and analytic
  functions.
\newblock \emph{Neural computation}, 8\penalty0 (1):\penalty0 164--177, 1996.

\bibitem[Montufar et~al.(2014)Montufar, Pascanu, Cho, and
  Bengio]{montufar_number_2014}
Guido~F Montufar, Razvan Pascanu, Kyunghyun Cho, and Yoshua Bengio.
\newblock On the {Number} of {Linear} {Regions} of {Deep} {Neural} {Networks}.
\newblock In \emph{Advances in {Neural} {Information} {Processing} {Systems}},
  volume~27. Curran Associates, Inc., 2014.
\newblock URL \url{https://arxiv.org/abs/1402.1869}.

\bibitem[Pascanu et~al.(2013)Pascanu, Montufar, and Bengio]{pascanu2013number}
Razvan Pascanu, Guido Montufar, and Yoshua Bengio.
\newblock On the number of response regions of deep feed forward networks with
  piece-wise linear activations.
\newblock \emph{arXiv preprint arXiv:1312.6098}, 2013.

\bibitem[Pinkus(1999)]{pinkus1999approximation}
Allan Pinkus.
\newblock Approximation theory of the mlp model in neural networks.
\newblock \emph{Acta numerica}, 8:\penalty0 143--195, 1999.

\bibitem[Poggio et~al.(2017)Poggio, Mhaskar, Rosasco, Miranda, and
  Liao]{poggio2017and}
Tomaso Poggio, Hrushikesh Mhaskar, Lorenzo Rosasco, Brando Miranda, and Qianli
  Liao.
\newblock Why and when can deep-but not shallow-networks avoid the curse of
  dimensionality: a review.
\newblock \emph{International Journal of Automation and Computing}, 14\penalty0
  (5):\penalty0 503--519, 2017.

\bibitem[Rolnick and Tegmark(2017)]{rolnick2017power}
David Rolnick and Max Tegmark.
\newblock The power of deeper networks for expressing natural functions.
\newblock \emph{arXiv preprint arXiv:1705.05502}, 2017.

\bibitem[Safran and Shamir(2017)]{safran2017depth}
Itay Safran and Ohad Shamir.
\newblock Depth-width tradeoffs in approximating natural functions with neural
  networks.
\newblock In \emph{International conference on machine learning}, pages
  2979--2987. PMLR, 2017.

\bibitem[Safran et~al.(2019)Safran, Eldan, and Shamir]{safran2019depth}
Itay Safran, Ronen Eldan, and Ohad Shamir.
\newblock Depth separations in neural networks: what is actually being
  separated?
\newblock In \emph{Conference on Learning Theory}, pages 2664--2666. PMLR,
  2019.

\bibitem[Telgarsky(2015)]{telgarsky2015representation}
Matus Telgarsky.
\newblock Representation benefits of deep feedforward networks.
\newblock \emph{arXiv preprint arXiv:1509.08101}, 2015.

\bibitem[Venturi et~al.(2021)Venturi, Jelassi, Ozuc, and Bruna]{osti_10329461}
Luca Venturi, Samy Jelassi, Tristan Ozuc, and Joan Bruna.
\newblock Depth separation beyond radial functions.
\newblock \emph{Journal of machine learning research}, 23\penalty0 (122), 2021.
\newblock URL \url{https://par.nsf.gov/biblio/10329461}.

\bibitem[Yarotsky(2017)]{yarotsky2017error}
Dmitry Yarotsky.
\newblock Error bounds for approximations with deep relu networks.
\newblock \emph{Neural Networks}, 94:\penalty0 103--114, 2017.

\bibitem[Zaslavsky(1975)]{zaslavsky1975facing}
Thomas Zaslavsky.
\newblock \emph{Facing up to arrangements: Face-count formulas for partitions
  of space by hyperplanes: Face-count formulas for partitions of space by
  hyperplanes}, volume 154.
\newblock American Mathematical Soc., 1975.

\end{thebibliography}
\newpage
\appendix

\section{Proof of Theorem \ref{mr2}}
\subsection{Exponential Network} \label{proof}

In this subsection, we derive explicit construction of $\exp$-network representing univariate polynomials, and use this construction to approximate the function $x \mapsto \ln x$. We remark that no bias term is needed for $\exp$-network as for every inner neuron:
\begin{align*}
\exp(\textbf{w}\textbf{x}+b)=\exp(b) \exp(\textbf{w}^T\textbf{x})
\end{align*}

\subsubsection{Univariate Polynomials}
In the next lemma we show that any polynomial can be approximated using $\exp$-network. Additionally, we bound the weights of the network in terms of $\varepsilon$ and the degree of the polynomial. Up to small changes, this result is proven in \cite{safran2019depth}, Lemma 3. 

\begin{lemma}
Consider the univariate monomial $p(x)=x^n$ defined over $K=[0,1]$. Then 
for every $\varepsilon > 0$
there exists $\hat{f} \in \M_{n+1}^1(\exp)$ that $\varepsilon$-approximates $p$. 
Moreover, 
$\hat{f} = \sum_{i=0}^n \nu_i \exp(w_i x)$ satisfies 
$|w_ix_i| \leq 1$ for every $x \in K $ and $|\nu_i| \leq  n! \varepsilon^{-\eta n}$ for some $\eta >0$.
\end{lemma}
\begin{proof}
Consider the function $w \in \RR \mapsto \exp(wx)$ and notice $\frac{d^n}{d w^n} \sigma(wx) \mid_{w=0} = x^n$. Representing the derivative as a Riemann derivative (e.g. \cite{ash1970characterization}, first page):
\begin{align*}
 \frac{d^n}{d w^n} \sigma(wx) \mid_{w=0} = \lim_{h \to 0}  h^{-n} \sum_{i=0}^n \binom{n}{i} (-1)^{n-i} \exp\bigg( (i-\frac{n}{2}) h x \bigg)
 \end{align*} 
 and the converges is uniform in $x$ over $K$. Hence, we can find $h = h(\varepsilon)$ small enough such that 
\begin{align*}
\hat{f}_\varepsilon(x) = h^{-n} \sum_{i=0}^n \binom{n}{i} (-1)^{n-i} \exp\bigg( (i-\frac{n}{2}) h x \bigg)
\end{align*} 
 is a shallow neural network with $n+1$ neurons that $\varepsilon$-approximates $p$, $|(i-\frac{n}{2}hx)| \leq 1$ and
 \begin{align*}
 |h|^{-n} \binom{n}{i} \leq n! \varepsilon^{-\eta n}
 \end{align*}
 for some $c>0$.
\end{proof}

\begin{corollary} \label{uni-poly}
Let $p: [0,1] \longrightarrow \RR$ be a $(n-1)$-degree polynomial defined by $p(x) = \sum_{i=0}^{n-1} a_i x^i$ with $|a_i| \leq \xi$. Then for every $\varepsilon > 0$, exists a shallow $\exp$-network $\hat{g} = \sum_{i=1}^{n^2} \nu_i \exp(w_i x) \in \M_m^1(\exp)$ with $m=n^2$ neurons 
such that:
\begin{enumerate}
\item $||p-\hat{g} || < \varepsilon$
\item $\max_{i \in [n^2], x \in [0,1]} |w_i x_i| \leq 1$ 
\item $|\nu_i| \leq (\frac{\xi \cdot n}{\varepsilon} )^{\OO(n)}$
\end{enumerate}
\end{corollary}

\subsubsection{The Logarithmic Function} 
In the next lemma we approximate the logarithmic function using the $\exp$ activation function. 
\begin{lemma} \label{lnx}
Let $\delta \in (0,1)$. The function $x \mapsto \ln(x)$ defined over $(\delta, 1]$ can be $\varepsilon$-approximated by $\hat{g} = \sum_{i=1}^n \nu_i \exp(w_i x) \in \M_{n}(\exp)$  with $n = \OO(\varepsilon^{-2}\delta^{-2})$, $|w_i x| \leq 1$ and $|\nu_i| = ( \frac{2}{\varepsilon^{-2} \delta^{-1}} )^{\OO(\varepsilon^{-1} \delta^{-1})}$
\end{lemma}
\begin{proof}
Defined $g_r(x)=\sum_{i=1}^r \frac{(-1)^{i+1}}{i} (x-1)^i$. By Taylor's theorem, for every $x \in (\delta, 1)$
exists $c \in (x, 1)$ such that:
\begin{align*}
\bigg| \ln x - \sum_{i=1}^r \frac{(-1)^{i+1}}{i} (x-1)^i \bigg| = \bigg|\frac{(\ln x)^{(i+1)}(c)}{(r+1)!} (1-x)^{r+1 } \bigg|= \frac{1}{r+1} c^{-i} (x-1)^{r+1} \leq \frac{1}{r+1} \frac{1}{\delta}
\end{align*}
choose $r = \ceil{2 \varepsilon^{-1} \delta^{-1} -1} $. By Corollary \ref{uni-poly}, there exists $\hat{g} \in \M_{n}(\exp)$ for $n=(r+1)^2=\OO(\varepsilon^{-2}\delta^{-2})$ with $||g_r - \hat{g}|| < \frac{\varepsilon}{2}$ on $K$. Finally for every $x \in (\delta,1]$:
\begin{align*}
| \ln x - \hat{g}(x) | \leq | \ln x -g_r(x)| + |g_r(x) - \hat{g}(x)| < \frac{\varepsilon}{2}+\frac{\varepsilon}{2} = \varepsilon
\end{align*}
\end{proof}
We remark that the $\OO(\varepsilon^{-2}\delta^{-2})$ can possibly be improved somewhat by taking an optimal polynomial approximation.

\subsection{Approximating The Product Function}
\subsubsection{Two-Hidden Layers}
\begin{lemma} \label{two-layers-appx}
Let $\varepsilon \in (0,1)$, and $C>0$. Denote $\delta := e^{-\frac{C}{d}}$. There exists 
\begin{align*}
f(\textbf{x}) = \exp \bigg( \sum_{i=1}^n \nu_i \exp(\textbf{w}_i \textbf{x}) \bigg)\in \M_n^2(\exp)
\end{align*}
 that satisfies the following:
\begin{enumerate}
\item $||p_d-f|| < \varepsilon$ on $K:=(\delta, 1)^d$
\item $n=\OO(\varepsilon^{-2}d^2)$
\item $|\textbf{w}_i \textbf{x}| < 1$ for any $x \in K$ and $i \in [n]$.
\item $|\nu_i| \leq ( \varepsilon^{-3} d^2)^{\OO(C \varepsilon^{-1})} $
\end{enumerate}
\end{lemma}
This result is proved in \cite{blanchard2021shallow} (Proposition 3.2) for ReLU networks. Up to minor details, the bellow construction follows along a similar line for the $\exp$ activation function while making sure $|\textbf{w}_i \textbf{x}| < 1$.
\begin{proof}
Consider the function $h: (\delta,1) \longrightarrow \RR$ defined by $h(x) = \ln x$. 
By Lemma \ref{lnx}, there exists $\hat{h} \in \M_{m}^1(\exp)$ with $m=M \varepsilon^{-2} d^2$ neurons with $||\hat{h} - h|| < \frac{\varepsilon}{d \cdot e} $, where $M$ is a constant independent of $d$. We can choose such $\hat{h}$ with the property that any $\textbf{w}_i$ weights vector in the network satisfies $|\textbf{w}_i \textbf{x}| < 1$.
Using $\hat{h}$, we can construct the following two-hidden layer network:
\begin{align*}
f(\textbf{x}) = \exp \bigg(\sum_{i=1}^d \hat{h}(x_i) \bigg) \in \M_m^2(\exp)
\end{align*}
Now let $\textbf{x} \in (\delta, 1)^d$.  Denote $\hat{y} = \sum_{i=1}^d \hat{h}(x_i)$ and $y = \sum_{i=1}^d \ln x_i$. Then by the mean value theorem there exists $c$ between $y$ and $\hat{y}$ with:
\begin{align*}
\begin{split}
| f(\textbf{x}) - p(\textbf{x}) | &= |\exp(y)- \exp(\hat{y})| = e^{c}  | y-\hat{y}| < e^{c} \frac{\varepsilon}{e} < \varepsilon
\end{split}
\end{align*}
where in the last step we use the fact that
since $y < 0$ and $|y-\hat{y}|<\varepsilon$, we have that $c< \varepsilon < 1$. It follows that $e^{c-1} < 1$.
\end{proof}
\begin{remark} \label{bounded_output}
In the proof of Lemma \ref{two-layers-appx}, notice that:
\begin{align*}
-C- \varepsilon < \sum_{i=1}^d \hat{h}(x_i) < \varepsilon
\end{align*}
in other words, the output of the second layer is bounded in an interval that does not depend on $d$.
\end{remark}
\subsubsection{Flattening To One-Hidden Layer}
Let $p_r$ be some $r$-degree univariate polynomial. In the next lemma we show that if we can replace the second layer activation function with $p_r$, we get a shallow 1-hidden layer network without adding much neurons:
\begin{lemma} \label{poly_on_shallow}
Let $p_r(x) = \sum_{i=0}^r a_i x^i$ a $r$-degree polynomial. Then: $p_r(\M_n^1(\exp)) \subset \M_{(n+1)^r}^1 (\exp)$.  That is, for any $f(\textbf{x}) = \sum_{j=1}^n \nu_j \exp(\textbf{w}_j^T\textbf{x}) \in \M_n^1(\exp)$, 
$p_r(f(\textbf{x}))$ is a shallow $\exp$-network:
\begin{align*}
p_r(f(\textbf{x})) = \sum_{i=1}^m \mu_i \exp((\textbf{w}_j')^T \textbf{x}_j)
\end{align*}
with $m \leq (n+1)^r$ neurons. Moreover:
\begin{enumerate}
\item if 
for every neuron $j \in [n]$, the input $\textbf{w}_j \textbf{x}$ satisfies 
$|\textbf{w}_j \textbf{x}_j | \leq a$, then the input for every neurons in $p_r(f(\textbf{x}))$ satisfies $|(\textbf{w}_j')^T \textbf{x}| \leq r \cdot a$.
\item If $|\nu_j| \leq N$ for every $j \in [n]$, then $|\mu_j| \leq \OO(N^r)$
\end{enumerate}

\end{lemma}
\begin{proof}
For every $0 \leq i \leq r$:
\begin{align*}
\begin{split}
(f(\textbf{x}))^i &= \bigg( \sum_{j=1}^n \nu_j \exp(\textbf{w}_j^T \textbf{x})   \bigg)^i = \sum_{k_1+...+k_n = i} \binom{i}{k_1,...,k_n} \prod_{j=1}^n \nu_j^{k_j} \exp(k_j \textbf{w}_j^T \textbf{x}) \\
&= \sum_{k_1+...+k_n = i} \binom{i}{k_1,...,k_n}  \bigg( \prod_{j=1}^n \nu_j^{k_j} \bigg) \exp\bigg( (\sum_{j=1}^n k_j \textbf{w}_j)^T\textbf{x} \bigg)
\end{split}
\end{align*}
hence:
\begin{align*}
\begin{split}
 p_r(f(\textbf{x})) &= \sum_{i=0}^r a_i \bigg( \sum_{j=1}^n \nu_j \exp(\textbf{w}_j^T\textbf{x}) \bigg)^i = \sum_{i=0}^r \sum_{k_1+...+k_n = i} \binom{i}{k_1,...,k_n}  \bigg( \prod_{j=1}^n \nu_j^{k_j} \bigg) \exp\bigg( (\sum_{j=1}^n k_j \textbf{w}_j)^T\textbf{x} \bigg)
 \end{split}
 \end{align*} 
 this is a shallow $\exp$ network with $m$ neurons for
 $ m =  \binom{n+r}{r} = \frac{1}{r!} (n+1)(n+2)\dots (n+r) = \prod_{i=1}^r (\frac{n}{i} + 1) \leq (n+1)^r$. Moreover, for any $\textbf{x}$:
\begin{align*}
\bigg| (\sum_{j=1}^n t_j \textbf{w}_j)^T\textbf{x} \bigg| \leq \sum_{j=1} t_j | \textbf{w}_j \textbf{x}| \leq r a
\end{align*}
\end{proof}

\begin{lemma}[\cite{safran2019depth} (Lemma 4), \cite{osti_10329461} (Lemma 30)]  \label{approx_exp}
Let $f: (a,b) \longrightarrow \RR$ be $L$-Lipschitz function. Then for any $\varepsilon > 0$, there exists a polynomial $p$ of degree $n = \ceil{4 (b-a) \varepsilon^{-3}L^3}$ such that $||  p - f|| < \varepsilon$. Moreover, $p(x) = \sum_{i=0}^n a_i x^i$ can be chosen such that $|a_i| \leq 2^n (b-a)^{1-i}$ and $|a_0| \leq 1+|f(0)|$
\end{lemma}
\begin{remark} \label{remark_on_exp}
In particular, the function $x \mapsto \exp(x)$ can be $\varepsilon$-approximated on $(-(C+1),1)$ by a polynomial of degree $\ceil{ 4(C+2) \varepsilon^{-3} e^3 }=\OO(\varepsilon^{-3})$.
\end{remark}

\begin{lemma}[The Flattening Lemma] \label{FlatteningLemma}
Let $f: (\delta,1)^d \longrightarrow \RR$ defined by $f(\textbf{x}) =\exp\bigg(  \sum_{i=1}^n a_i \exp(\textbf{w}_i^T \textbf{x})  \bigg) \in \M_n^2(\exp)$. 
Assume that for every $\textbf{x} \in (\delta, 1)^d$ we have $\sum_{i=1}^n a_i \exp(\textbf{w}_i^T \textbf{x}) \in (-k,1)$. Then
for every $\varepsilon > 0$ there exists $g \in \M_m^1(\exp)$ with $m=(n+1)^{\OO(\varepsilon^{-3})}$ such that $||g-f|| < \varepsilon$. 
\end{lemma}
\begin{proof}
Let $\varepsilon > 0$. Let $p_r$ given by Lemma \ref{approx_exp} and Remark \ref{remark_on_exp} such that $| e^x - p_r(x) | < \varepsilon$ for every $x \in (-k,1)$, where $p_r$ is a polynomial of degree $r = \OO(\varepsilon^{-3})$.
 By Lemma \ref{poly_on_shallow}, $g = p_r(\sum_{i=1}^n a_i \exp(\textbf{w}_i \textbf{x}))$ is a $\exp$-shallow network with $(n+1)^r$ neurons. Denote $y= \sum_{i=1}^n a_i \exp(\textbf{w}_i^T \textbf{x}) \in [a,b]$, then:
\begin{align*}
| f(\textbf{x}) - g(\textbf{x})| = | \exp(y) - p_r(y) | < \varepsilon
\end{align*}
\end{proof}

\begin{lemma}
Let $C>0$, $d \in \N$ and $p: (\delta:= e^{\frac{-C}{d}},1)^d \longrightarrow [0,1]$ defined by $p(\textbf{x}) = \prod_{i=1}^d x_i$. For any $\varepsilon \in (0,1)$, there exists a shallow $\exp$-network $f_n =\sum_{i=1}^n b_i\exp(\textbf{w}_i \textbf{x})\in \M_n^1(\exp)$ with:
\begin{enumerate}
\item $|| p-f_n|| < \varepsilon$ 
\item  $n= (M \varepsilon^{-2} d^{2})^{\OO(\varepsilon^{-3})}$, where $M$ is a constant independent of $\varepsilon$ and $d$.
\item for any $\textbf{x} \in (\delta, 1)^d$, and $i \in [n]$, $\nu_i\exp(\textbf{w}_i \textbf{x}) = \OO(\exp(C\varepsilon^{-1}))$
\end{enumerate}
\end{lemma}

\begin{proof}
By Lemma \ref{two-layers-appx}, there exits a two-layer $\exp$-network $g_n$:
\begin{align*}
g_m(\textbf{x}) = \exp\bigg( \sum_{i=1}^m a_i \exp((\textbf{w}'_i )^T\textbf{x}) \bigg)
\end{align*}
 with $m=M \varepsilon^{-2} d^2$ neurons on the first layer and one neuron on the second such that $|| p - g_n|| < \frac{\varepsilon}{2}$ and $M$ is a constant that depend only on $C$. Additionally, by Remark \ref{bounded_output}, it holds that for every $\textbf{x} \in (\delta,1)^d$:
 \begin{align*}
  \sum_{i=1}^n a_i \exp(\textbf{w}_i \textbf{x}) \in (-(C-1), 1)
 \end{align*}
 Hence by Lemma \ref{FlatteningLemma}, there exists a shallow $\exp$-network $f_n$ with $n = (M \varepsilon^{-2} d^{2})^{\OO(\varepsilon^{-3})}$ such that $||f_n - g_m|| < \frac{\varepsilon}{2}$. Finally:
 \begin{align*}
 |p(\textbf{x}) - f_n(\textbf{x}) | \leq |p(\textbf{x}) - g_m(\textbf{x}) | + |g_m(\textbf{x}) - f_n(\textbf{x})| < \varepsilon
 \end{align*}
\end{proof}

\subsection{ReLU Networks}
\begin{lemma}[\cite{blanchard2021shallow} (Lemma B.2)] \label{exp2relu}
Let $f: I \longrightarrow [a,b]$ be a continuous increasing or decreasing function where $I$ is an interval and let $\varepsilon > 0$. There exists a ReLU-shallow network with $\ceil{\frac{b-a}{\varepsilon}}$ neurons that $\varepsilon$-approximates $f$ 
\end{lemma}
Therefore we can approximates
 the exponential function $\nu e^x: I \longrightarrow [a,b]$ with $\ceil{(b-a) \varepsilon^{-1}}$  neurons for any $\nu \in \RR$.

\begin{lemma}[Shallow $\exp$ to shallow ReLU]
Let $f(\textbf{x}) = \sum_{i=1}^n \nu_i \exp(\textbf{w}_i \textbf{x})$ be a shallow $\exp$-neural network defined over $K$. Denote $C_1 := \max_{i \in [n], \textbf{x} \in K} |\textbf{w}_i \textbf{x}|$ and $C_2 := \max_{i \in [n], \textbf{x} \in K} |\nu|_i \exp(\textbf{w}_i \textbf{x})$. Then 
for every $\varepsilon > 0$, 
exists a ReLU network $\hat{f}$ with $||g-\hat{f}|| < \varepsilon$ and $\hat{f}$ has $\ceil{2n^2 C_2 \varepsilon^{-1}}$ neurons.
\end{lemma}

\begin{proof}
Let $\varepsilon > 0$. For every $i \in [n]$, consider the function $f_i:[-C_1, C_1] \longrightarrow [-C_2, C_2]$ defined by $f_i(y) = \nu_i e^y$.
Invoke Lemma \ref{exp2relu} to obtain $\hat{f}_i \in \M_m^1(\text{ReLU})$ with $||f_i - \hat{f}_i|| < \frac{\varepsilon}{n}$ and $m = \ceil{2nC_2 \varepsilon^{-1}}$. The shallow ReLU network $\hat{f}(\textbf{x}) := \sum_{i=1}^n \hat{f}_i(\textbf{w}_i \textbf{x})$ has $\ceil{2n^2 C_2 \varepsilon^{-1}}$ neurons and $\varepsilon$-approximates $f$.
\end{proof}

\newpage
\section{Exact Approximation}   \label{new_proof_to_Lin}
\begin{theorem}[\cite{lin2017does}]
Let $\sigma \in C^\infty(\RR)$ be a smooth non-polynomial activation function and $K=[-k,k]^d$. Then for every multivariate monomial $p(x) = \prod_{i=1}^d x_i$ defined over $K$ and every $\varepsilon>0$,
there exists $f_n \in \M_n^1(\sigma)$ with $n=2^d$ such that  $\varepsilon$-approximates $p$. In other words, $n$ is independent of $k$.
\end{theorem}

This statement was proven by \cite{lin2017does}. For completeness, we present a slightly different proof, utilizing arguments from 
\cite{leshno1993multilayer}, \cite{mhaskar1996neural}. Notice that in contrast to 
ReLU networks,
$n$ is independent of $k$, and in-fact independent of $\varepsilon$ as well. For a smooth activation, the weights, but not the number of neurons, may not dependent on the domain or the approximation error.

\begin{proof}
Since $\sigma \in C^\infty$ and not a polynomial, there exists a point $x_0$ for which $\sigma^{(k)} (x_0) \neq 0$ for any $k \in \N$ (e.g. \cite{pinkus1999approximation} and \cite{corominas1954condiciones}). Consider the output of a neuron as a function of the weights: $\sigma(w;x)=\sigma(w^T x +x_0)$. It follows that:
\begin{align}\label{partial_der}
\frac{\partial^n}{\partial w_1...\partial w_d} \sigma(w^T x +x_0) = \prod_{i=1}^d x_i \sigma^{(n)}(w^T x + x_0)
\end{align}
in particular for $w=0_{\RR^d}$:
\begin{align*}
\prod_{i=1}^d x_i = \frac{1}{\sigma^{(n)}(x_0) } \frac{\partial^n}{\partial w_1...\partial w_d} \sigma(w^T x + x_0)  \mid_{w=0}
\end{align*}
Hence it suffices to find a shallow network $f_n \in \M_n^1(\sigma)$ that approximates the cross derivative of $\sigma(w;x)$ at $w=0$. 
\begin{align} \label{partial_diff}
\frac{\partial^n}{\partial w_1...\partial w_d} \sigma(w^T x + x_0)  \mid_{w=0} = \lim_{h \to 0^+}
\sum_{\textbf{s} \in \{0,1 \}^d} \frac{(-1)^{|\textbf{s}|} }{h^d} \sigma(h \cdot (2\textbf{s}-\textbf{1})^T \textbf{x} + x_0)
\end{align}
for a fix $h>0$, the right-hand side equation (\ref{partial_diff}) is a shallow neural network with $2^d$ neurons. Each neuron, up to scaling by $h$ and transitioning by $x_0$, corresponds to a subset $s \subset [d]$ of indices. Each feature $x_i$ with $i \in s$ receive a weight of $1$ and all other weight of $-1$. Since $x_i \leq k$ for every $i \in [d]$, we may choose $h$ small enough such that for every $\textbf{x} \in K$:
\begin{align}
\bigg|\frac{\partial^n}{\partial w_1...\partial w_d} \sigma(w^T x + x_0)  \mid_{w=0}  - \sum_{\textbf{s} \in \{0,1 \}^d} \frac{(-1)^{|\textbf{s}|} }{h^d} \sigma(h \cdot (2\textbf{s}-\textbf{1})^T \textbf{x} + x_0) \bigg| < \varepsilon \sigma^{(n)}(x_0)
\end{align}

We can therefore find $f \in \M_{2^d}(\sigma)$ with $|| f -  p || < \varepsilon$, as desired.
\end{proof}
%
%
%
%
%

\end{document}